%% file: main.tex
\documentclass[letterpaper, 10 pt, conference]{ieeeconf}
\IEEEoverridecommandlockouts
\usepackage[utf8]{inputenc}
\usepackage{amsmath,amssymb,amsfonts}
\usepackage{bm}
\usepackage[hyphens]{url}
\usepackage{hyperref}
\usepackage{graphicx,graphics}
\usepackage{multirow}
\usepackage{booktabs}
\usepackage[ruled,linesnumbered]{algorithm2e}
\usepackage{textcomp}
\usepackage{tabularx}
\usepackage{relsize}
\usepackage{soulutf8}
\usepackage[flushleft]{threeparttable}
\usepackage{cite}
\usepackage[table,dvipsnames]{xcolor} 
\usepackage{rotating}
\usepackage{pifont}
\newcommand{\midsepremove}{\aboverulesep = 0.2mm \belowrulesep = 0.2mm}
    \midsepremove
\newcommand{\midsepdefault}{\aboverulesep = 0.605mm \belowrulesep = 0.984mm}
    \midsepdefault
\newcommand{\mytick}{\textcolor{Green}{\small{\ding{52}}}}
\newcommand{\mytickb}{\textcolor{Plum}{\small{\ding{52}}}}

\graphicspath{{figures/}}
\DeclareGraphicsExtensions{.pdf,.jpeg,.png,.tiff,.jpg}
\hypersetup{
    colorlinks=true,
    linkcolor=black,    
    urlcolor=black,
}
\title{\LARGE \bf SpectGRASP: Robotic Grasping by Spectral Correlation}
\author{{Maxime Adjigble$^{\dagger}$, Cristiana de Farias, Rustam Stolkin, Naresh Marturi}
\thanks{All authors are with the Extreme Robotics Laboratory, School of Metallurgy and Materials, University of Birmingham, Edgbaston, B15 2TT, UK. $^{\dagger}$Corresponding author email: {\tt\small m.k.j.adjigble@bham.ac.uk}}%
\thanks{This work was supported by the UK National Centre for Nuclear Robotics, part-funded by EPSRC EP/R02572X/1 and in part supported by CHIST-ERA under Project EP/S032428/1 PeGRoGAM.}%
}
\begin{document} \sloppy
\bstctlcite{IEEEexample:BSTcontrol}
\maketitle%
\input{sections/abstract.tex}
%
\section{Introduction}
\label{sec:intro}
\input{sections/introduction.tex}
%
\section{Methodology}
\label{sect:method}
\input{sections/method.tex}
\section{Experimental Validations}
\label{sect:results}
\input{sections/results.tex}
\section{Conclusion}
\label{sect:conc}
\input{sections/conclusion.tex}
%
%
\bibliographystyle{IEEEtran}
\bibliography{IEEEabrv,references}
%
\end{document}

%% file: sections/abstract.tex
\begin{abstract}
This paper presents a spectral correlation-based method (SpectGRASP) for robotic grasping of arbitrarily shaped, unknown objects. Given a point cloud of an object, SpectGRASP 
extracts contact points on the object's surface matching the hand configuration. It neither requires offline training nor a-priori object models. 
We propose a novel Binary Extended Gaussian Image (BEGI), which represents the point cloud surface normals of both object and robot fingers as signals on a 2-sphere. Spherical harmonics are then used to estimate the correlation between fingers and object BEGIs.
The resulting spectral correlation density function provides a similarity measure of gripper and object surface normals. This is highly efficient in that it is simultaneously evaluated at all possible finger rotations in $\bm{\mathrm{SO(3)}}$. 
A set of contact points are then extracted for each finger using rotations with high correlation values. We then use our previous work, Local Contact Moment (LoCoMo) similarity metric, to sequentially rank the generated grasps such that the one with maximum likelihood is executed. %
We evaluate the performance of SpectGRASP by conducting experiments with a 7-axis robot fitted with a parallel-jaw gripper, in a physics simulation environment.
Obtained results indicate that the method not only can grasp individual objects, but also can successfully clear randomly organized groups of objects. The SpectGRASP method also outperforms the closest state-of-the-art method in terms of grasp generation time and grasp-efficiency.
%





\end{abstract}

%% file: sections/introduction.tex
Reliable autonomous grasping of arbitrarily shaped, unknown objects, in cluttered and unstructured environments, remains a challenging problem. Major difficulties lie in establishing a useful correspondence, between a robot's gripper fingers and an object's surface obtained from computer vision, to achieve a robust and stable grasp. Although, robots have already been deployed in flexible manufacturing to perform vision-guided pick and place tasks, 
these predominantly assume that precise object models are known a-priori for each product, for which stable grasp poses can be pre-computed. For robots to reliably manipulate arbitrarily shaped objects, in unstructured environments, robust and efficient methods for grasp synthesis are needed. In this context, given a gripper configuration, we present a novel spectral domain method that can generate grasps for unknown objects observed as point clouds.
\begin{figure}
    \centering
    \includegraphics[width=\columnwidth]{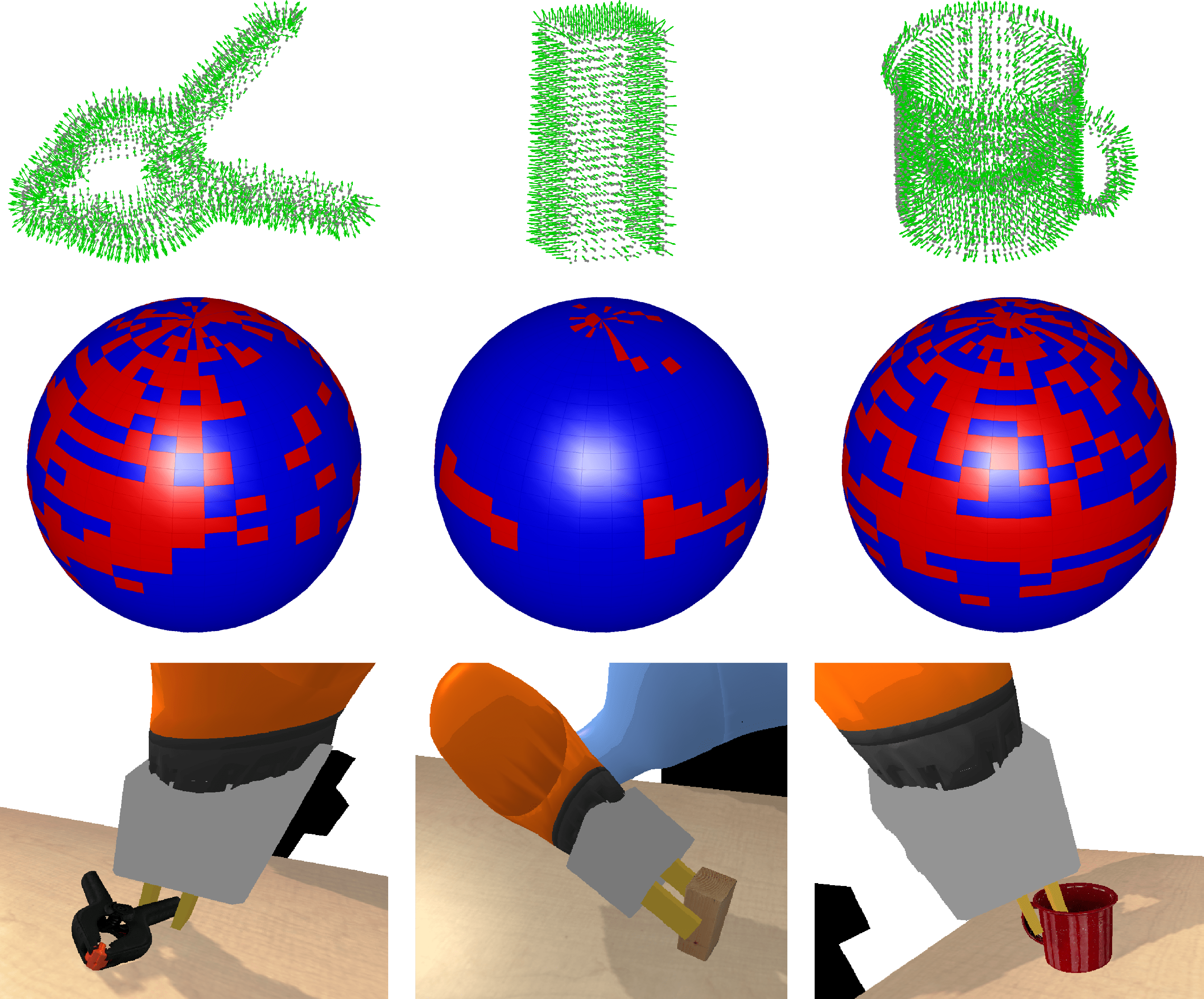}
    \caption{Object grasping by proposed SpectGRASP method. (Top) Perceived object point clouds with surface normals. (Middle) Corresponding Binary Extended Gaussian Images. (Bottom) Executed best grasps.}
    \label{fig:teaserimage}
\end{figure}

A variety of methods have emerged in the literature for robotic grasping \cite{bicchi2000robotic, bohg2013data, shimoga1996robot, sahbani2012overview}. Most of the state-of-the-art methods can be classified into analytical approaches versus data-driven methods \cite{sahbani2012overview}. The conventional analytical approaches predominantly generate grasp hypotheses by using physics-based formulations. They largely rely on a-priori information like accurate geometric models of objects, dynamic parameters such as object mass, and assumed values for friction coefficients etc. These methods achieve acceptable performance when grasping at prescribed contact points. However, they have difficulty in generalising to many real-world scenarios, particularly for unknown objects, of arbitrary shape, in unstructured scenes. These problems have been addressed by more recent data-driven methods, which sample grasp candidates by means of \textit{learning} \cite{ten2017grasp,marturi2019dynamic,levine2018learning,mahler2017dex,kopicki2016one}. These methods enabled robots to better cope with uncertainties in vision and perception, and demonstrated some degree of generalisation to new objects. However, they predominantly require large amounts of data and extensive offline training. Moreover, their ability to handle novel objects, especially those with shapes that differ greatly from training data objects, is limited. Also, grasping objects from randomly organised cluttered scenes, remains a major problem to solve. These limitations can be tackled by using geometry-based approaches, like the one presented in this paper.%

Although geometric methods may not be as fast as some learning-based methods, they are: easy to generalise for multi-fingered hands; work well for grasping novel objects with a-priori unknown geometries; and can handle heaps of objects in cluttered and unstructured environments\cite{adjigble2018model}. Recent methods, based on shape similarities between hand fingers and object surfaces, have demonstrated remarkable results for robotic grasping of unknown objects \cite{kiatos2020geometric,sorour2019grasping,adjigble2018model,eppner2013grasping,charusta2012independent}. These methods synthesise grasp hypotheses by matching the surface regions of gripper fingers with similar surface patches on the object, to maximize the contact area during grasping. The sampled grasp candidates are then ranked or sorted by virtue of custom grasp ranking metrics to find the candidate with highest probability of success. Our previous work in this area presented a Local Contact Moment (LoCoMo) similarity metric-based grasp planner\cite{adjigble2018model}. This method demonstrated high success rates in grasping individual unknown objects and also in clearing random cluttered heaps of objects. However, it requires high computational time (about ten seconds), which also increases with the point cloud density. A comparable approach based on shape adaptability has been presented in \cite{eppner2013grasping}. Very recently, an approach based on gripper workspace spheres was presented in \cite{sorour2019grasping}, for grasping unknown objects represented as registered point clouds. This approach can be flexibly generalised to a variety of grippers. Although the method is comparably fast, it can only be applied to scenes containing single objects. A fast geometric unknown object grasping algorithm based on force balanced optimisation was presented in \cite{lei2014fast}. Like other methods, it also synthesises grasps based on online processing of a single point cloud. In contrast to the aforementioned approaches, there are other geometric methods that work by surface reconstruction and object modelling \cite{de2021simultaneous,lippiello2012visual,quispe2015exploiting,vezzani2017grasping}. However, sampling grasp candidates for large numbers of object points and gripper configurations can be slow and computationally expensive.

To overcome these limitations, we propose SpectGRASP, a robust and computationally efficient method for synthesizing grasp candidates for unknown objects. At the core of our method, spherical harmonics are used to efficiently compute spectral correlations between object and finger surface normals. 
Previously, a method using spherical harmonics for grasping was reported in \cite{pokorny2014grasp}. It transfers a known grasp from one object to another by smoothly deforming the surface of the first object into the second one. Unlike our method, it requires a complete, dense point cloud, sampled from a closed convex surface, \textit{i.e.}, without holes. 
In contrast, SpectGRASP makes no assumptions about object shape, and is effective with sparse point clouds. Furthermore, while \cite{pokorny2014grasp} has only been demonstrated on single and mostly convex objects, SpectGRASP generates grasps for a wide range of object shapes, and robustly handles multiple objects in cluttered scenes.

SpectGRASP is designed as a two step process: correspondence matching and contact ranking. The matching is performed by correlating gripper and object surface normals. We propose Binary Extended Gaussian Images (BEGI) to represent surface normals.  A Fast Fourier Transform (FFT) for functions defined on the rotation group is used to efficiently compute the correlation density function \cite{kostelec2008ffts}. %
In the same way that cross-correlation is used as a similarity measure, between two signals over a set of positions in the standard Euclidean space $\mathbb{R}^n$, the resulting spectral correlation density function can be used as a similarity measure over $\bm{\mathrm{SO}(3)}$. Thus, finding a pattern embedded in a signal (both defined on the unit sphere) is reduced to finding pattern rotations with high correlation values. Here, the finger BEGI is the pattern and the object BEGI (seen in Fig. \ref{fig:teaserimage}) is the signal. From this correlation, a set of contact points are extracted on the object for each gripper finger. Finally, gripper poses are sampled from the contact points and ranked according to their LoCoMo score, which provides a grasp likelihood metric based on the correspondence between surface patches of gripper fingers and object surfaces \cite{adjigble2018model}. %
%
The main contributions of this work are as follows:
\begin{itemize}
    \item[(i)] We propose BEGI as a novel way of representing point cloud surface normals. This representation helps finding corresponding point-pairs between two point clouds.
    \item[(ii)] We propose the use of spectral correlation for efficiently sampling grasp contact points on the surface of an object given a robot hand configuration. 
    \item[(iii)] We present a new, generic and efficient grasp generation pipeline, combining spectral correlation with the LoCoMo grasp likelihood metric of our previous work. 
\end{itemize}
%
SpectGRASP works directly using online captured point clouds of objects. It does not require any object CAD models, or learning with offline training data. In simulation experiments, using a 7-axis robot with a two-finger hand, we demonstrate how SpectGRASP generates effective grasps for a variety of objects selected from the YCB object set \cite{calli2015ycb}. We also demonstrate how SpectGRASP can reliably clear randomly generated scenes of multiple objects with successive grasps. In addition, we compare and discuss the performance of SpectGRASP against the LoCoMo-based grasp planner \cite{adjigble2018model}, since this was the previous state-of-the-art in learning-free and model-free grasping.

%% file: sections/method.tex
In this section, we present our proposed SpectGRASP method. We first show how point cloud surface normals are represented as BEGI. Later, we show how spherical harmonics are used to compute correspondence between two BEGIs and estimate the grasp contact points. Finally, we present how the grasp candidates are generated and ranked.
\subsection{Binary Extended Gaussian Image (BEGI)}
\begin{figure}
    \centering
    \includegraphics[width=\columnwidth]{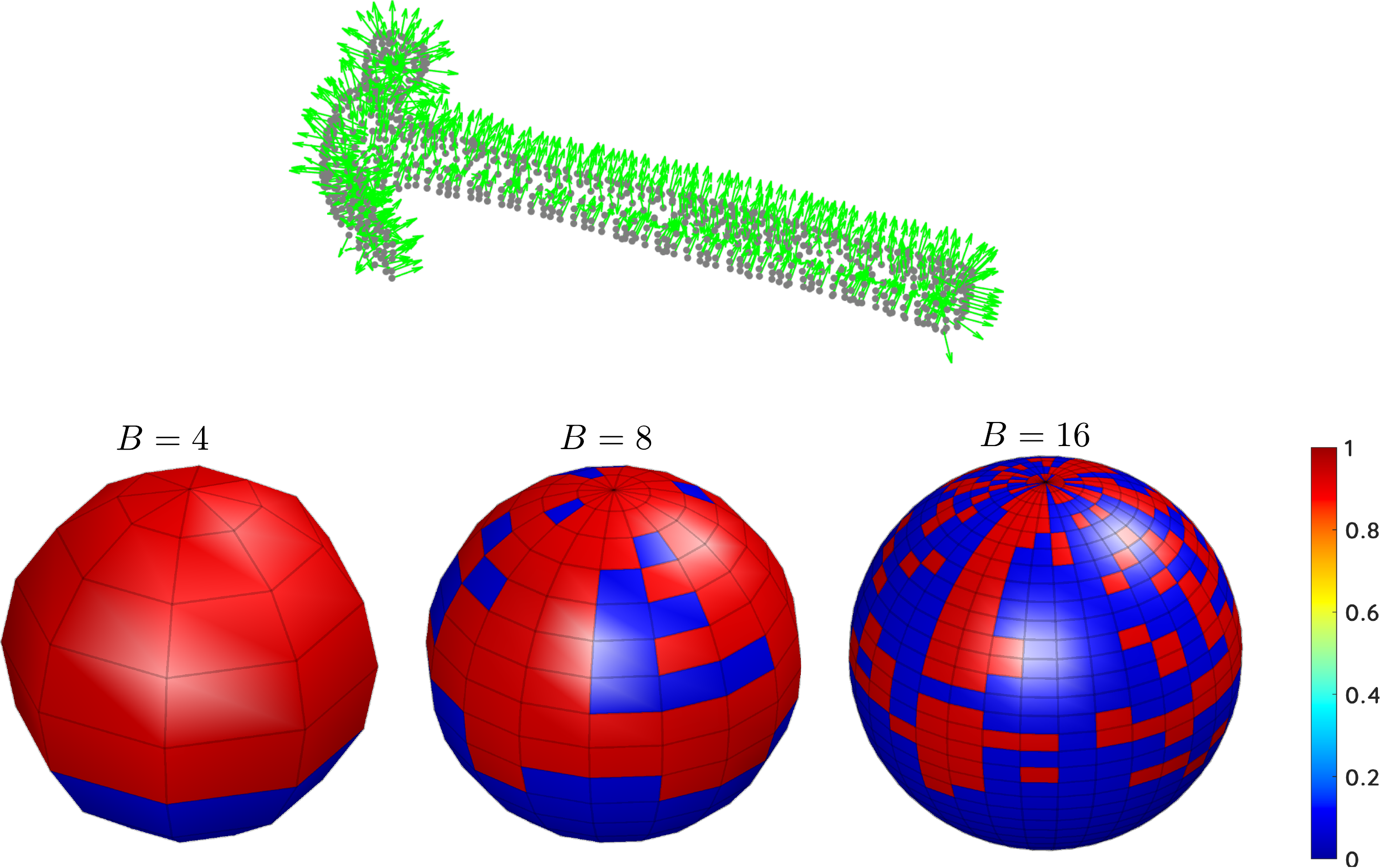}
    \caption{BEGIs with different bandwidths for the above ``Hammer'' object point cloud. The one with highest bandwidth, \textit{i.e.}, $B=16$, provides finer distribution where as the lower bandwidth ones show coarse distributions.}
    \label{fig:BEGI_bands}
\end{figure}

The extended Gaussian image (EGI) \cite{horn1984extended} has been extensively used as a shape descriptor for object surface normals \cite{lowekamp2002exploring,nayar1990specular,little1985extended}. It represents the distribution of surface normals in polar coordinate system by mapping them onto a unit sphere, also denoted as the 2-sphere or $\bm{S}^2$. Given a vector $\bm{n}=(n_x,n_y,n_z)$ in the Euclidean space  $\mathbb{R}^3$, the corresponding spherical coordinates $(r, \theta, \phi)$ are computed as:
\begin{equation}\label{eq:spherical_coordinates}
\begin{gathered}
    r = \sqrt{n_x^2 + n_y^2 + n_z^2} \quad \quad \theta = \arctan{\frac{\sqrt{n_x^2 + n_y^2}}{n_z}} \\
   \mkern-38mu \phi = \arctan(\frac{n_y}{n_x}) 
\end{gathered}
\end{equation}
Physically, these coordinates represent norm, longitude and latitude. When $\bm{n}$ is unitary, \textit{i.e.}, $r = 1$, the set $(\theta, \phi)$ is sufficient to locate the vector on the unit sphere. To make use of spherical coordinates to numerically compute distributions, a discrete sphere is required. We use the following discretization along longitude and latitude to represent object surface normals on $\bm{S}^2$: $\theta_j = \frac{\pi(2j + 1)}{4B}$ and $\phi_k = \frac{\pi k}{B}$  , $(j, k) \in \mathbb{N}$ with the constraint $0  \leq j, k < 2B$. Here, $B$ is the bandwidth and acts like a low pass filter. Smaller $B$ values lead to a coarser distributions and higher values provide more accurate representations, as seen in Fig. \ref{fig:BEGI_bands}. While conventional EGIs use histograms to represent the distribution of surface normals on a unit sphere, our method uses binary values for each cell parameterised by the indices $(j,k)$. This corresponds to the existence of at least one surface normal in the direction of $(\theta_j, \phi_k)$. Additionally, we store the list of point coordinates of surface normals corresponding to the cell. 

Fig. \ref{fig:BEGI_bands} illustrates object surface normals represented as BEGI for three different bandwidths. Finer representations on the sphere are obtained with higher bandwidth, \textit{i.e.}, $B=16$. Points belonging to adjacent cells when $B = 8$ are merged into a single cell at a lower bandwidth, $B = 4$. 
Given a point cloud with surface normals $\mathcal{PC}_n = \{(\bm{p},\bm{n}) \in  \mathbb{R}^3\times \bm{S}^2 \}$ and a bandwidth $B$,  the BEGI can be constructed using \eqref{eq:begi_representation}.
\begin{equation}\label{eq:begi_representation}
\begin{gathered}
    f(\theta_j, \phi_k)  = v_{jk} \\
   \mathcal{P}_{jk} = \left\{ p_i \in  \mathbb{R}^3 \mid n_i = (\theta_j, \phi_k) \right\}
\end{gathered}
\end{equation}
With $0  \leq j, k < 2B$,  $v_{jk} \in [0, 1]$. $v_{jk} = 1$ if there exists a surface normal $n_i =(\theta_j, \phi_k)$ in $\mathcal{PC}_n$ and $v_{jk} = 0$, otherwise. $f(\theta_j, \phi_k)$is the BEGI function and $\mathcal{P}_{jk}$ is the BEGI point set at $(j,k)$. Any point cloud with surface normals can be reconstructed upto a degree of precision using its BEGI by specifying an appropriate value for the bandwidth $B$.
\subsection{Spherical Harmonics}
Spherical harmonics are used to provide a decomposition of functions defined on the unit sphere \cite{muller2006spherical}. They are the spherical equivalent of the Fourier transform in Euclidean space and form an orthonormal basis on the sphere. In computer vision, they have been used for 3D model recognition and visual servoing \cite{cohen2018spherical,marturi2016image}. Any square-integrable function  $f :\bm{S}^2 \rightarrow \mathbb{C} $  ($f \in L^2(\bm{S}^2)$) can be represented by its spherical harmonic expansion as:
\begin{equation}
f(\theta, \phi) = \sum_{l=0}^{l_{max}}\sum_{m=-l}^{l}\hat{f_l^m}Y_l^m(\theta, \phi)
\label{eq:spherical_harmonics}
\end{equation}
where,  $l, m \in \mathbb{N}^+$ , $l_{max} > 0$ is the maximum degree of expansion,  $Y_l^m$ is the spherical harmonics of degree $l$ and order $m$, $\hat{f_l^m}$ is the corresponding harmonic coefficient. The spherical harmonics $Y_l^m$ are computed as:
\begin{equation}
Y_l^m(\theta, \phi) = (-1)^m\sqrt{\frac{(2l+1)(l-m)!}{4\pi(l+m)!}}P_l^m(cos\theta)e^{im\phi}
\label{eq:spherical_harmonics_yml}
\end{equation}
In the literature, the factor $(-1)^m$ is sometimes integrated with the definition of Legendre polynomials, $P_l^m$. 
The harmonic coefficients $\hat{f_l^m}$ are computed by the inner product between $f$ and $Y_l^m$ over $\bm{S}^2$:
\begin{equation}\label{eq:harmonic_coeffs}
\begin{aligned}
\hat{f_l^m} &= \int_{w \in \bm{S}^2} f(w)\overline{Y_l^m(w)} \,dw \\
&= \int_0^{2\pi}d\phi\int_0^{\pi}d\theta\sin\theta f(\theta, \phi)\overline{Y_l^m(\theta,\phi)}
\end{aligned}
\end{equation}
where, $\overline{Y_l^m}$ is the complex conjugate of  $Y_l^m$. Using \eqref{eq:spherical_harmonics}, the harmonic decomposition of a BEGI function $f : \bm{S}^2 \rightarrow [0,1] \subset \mathbb{C}$ is computed. This means that the harmonic functions can be used to evaluate the correlation of two BEGIs. 
The harmonic coefficients described in \eqref{eq:harmonic_coeffs} are computed numerically by using the previously presented discretization of the sphere. By setting a value for $B$, the angles $\theta$ and $\phi$ can be sampled using an equiangular  $2B\times2B$ grid.
\subsection{Spectral Correlation of Functions on \texorpdfstring{$\bm{\mathrm{SO}(3)}$}{}}
\begin{figure*}
    \centering
    \includegraphics[width=\textwidth]{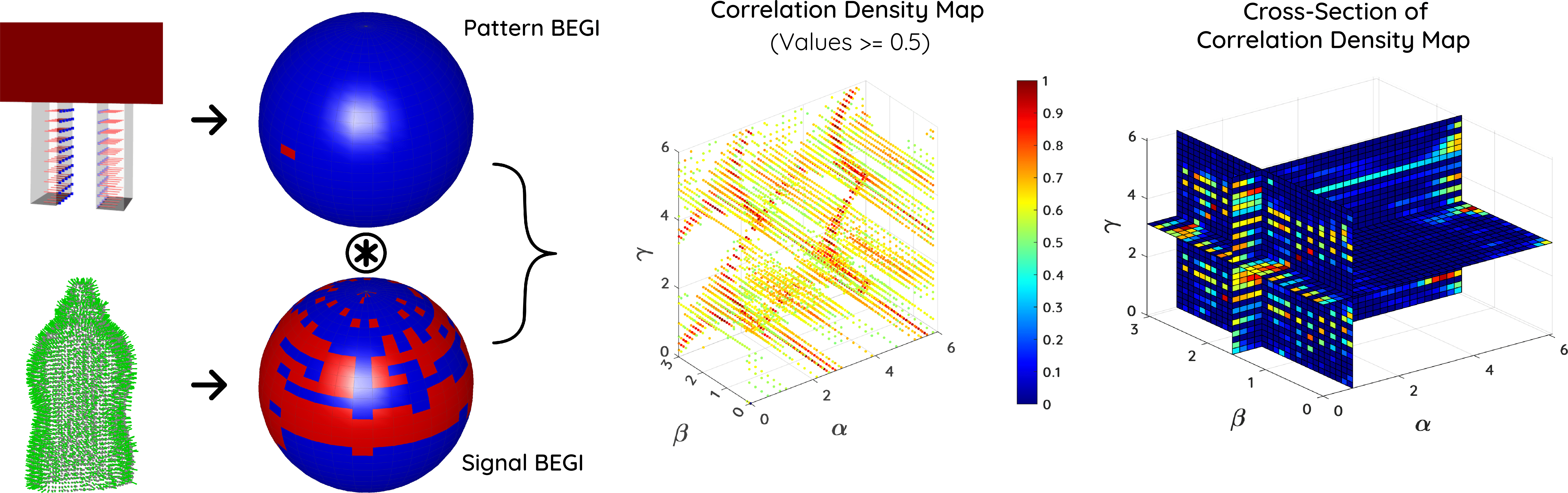}
    \caption{Illustation of BEGI correlation process. Initially, surface normals of both gripper fingers and object are converted to BEGIs. These are then correlated using spherical hormonics resulting in a correlation density map. Here we show the values $\geq 0.5$. The cross-section along $(\alpha, \beta, \gamma) = (\pi/4, \pi/2, \pi)$ shows an in depth view of the correlation density map.  Blue dots correspond to rotations with low correlation values, while red dots are locations with high correlation.
    }
    \label{fig:begi_correlation}
\end{figure*}

%
Let $f$ and $g$ are two functions with bandwidth $B$ defined on $\bm{S}^2$, and $g_r = g(\mathcal{R}(\alpha, \beta, \gamma)) = g(\alpha, \beta, \gamma)$ is the rotated version of $g$ by a rotation $\mathcal{R} \in \bm{\mathrm{SO}(3)}$\footnote{$\mathcal{R}$ is taken as $zyz$ Euler angles $\alpha, \beta, \gamma$; $\alpha, \gamma \in [0, 2\pi[$ and $\beta \in [0, \pi]$.}. The correlation $\mathcal{C}(\mathcal{R})$ between $f$ and $g_r$ is computed as
\begin{equation}
    \mathcal{C}(\mathcal{R}) = \int_{w \in S^2} f(w)\overline{g_r(w)} \,dw
    \label{eq:correlation}
\end{equation}
\eqref{eq:correlation} evaluates the degree of similarity between $f$ and $g_r$. By computing the correlation for a set of rotations, a correlation density function is obtained as shown in Fig. \ref{fig:begi_correlation}(d). 
In harmonic domain, rotations are expressed as \textit{Wigner D-Matrices}, $D_{mm'}^l$ \cite{wigner2012group}. %
Consequently, the rotated pattern BEGI $g_r$ is represented in the spectral domain as:
%
\begin{equation} \label{eq:wigner_D_matrix}
\begin{gathered}
        g(\alpha, \beta, \gamma) = \sum_{l=0}^{l_{max}} \sum_{m=-l}^{l} \sum_{m'=-l}^{l} D_{mm'}^l(\alpha, \beta, \gamma)\hat{g_l^{m'}}Y_l^m \\
        \mbox{with}\quad D_{mm'}^l(\alpha, \beta, \gamma) = e^{-im\alpha}d_{mm'}^l(\beta)e^{-im'\gamma}
\end{gathered}
\end{equation}
where, $d_{mm'}^l$ are the \textit{Wigner d-functions}. Replacing $f$ and $g_r$ in \eqref{eq:correlation} by their harmonic expansions, 
using the orthogonality principle of the harmonic coefficients, \textit{i.e.}, $\int_{w \in S^2} Y_l^m(w)\overline{Y_{l'}^{m'}(w)} \,dw  = 1$ if $m=m'$ and $0$, otherwise, we get:
\begin{equation}
    \mathcal{C}(\mathcal{R}) = \sum_{l=0}^{l_{max}}\sum_{m=-l}^{l}\sum_{m'=-l}^{l} \hat{f_l^m}\overline{\hat{g_l^{m'}}} \overline{D_{mm'}^l(\mathcal{R})} 
    \label{eq:correlation_simplified}
\end{equation}
\eqref{eq:correlation_simplified} can be evaluated using FFTs defined on $\bm{\mathrm{SO}(3)}$. We refer the readers to \cite{kostelec2008ffts} for more details on spectral correlation.
\subsection{Contact Sampling by Spectral Correlation}
A grasp is defined by the location of each finger of the gripper on an object and a pose defining the wrist location in space. For a gripper with $N_f$ fingers, the grasp configuration $\mathcal{G}$ can be defined as:
\begin{equation}
    \mathcal{G} = (p_1\cdots p_{N_f}, n_1\cdots n_{N_f}, M)
    \label{eq:grasp_conf}
\end{equation}
where, $p_i, n_i \in \mathbb{R}^3$ are respectively the object contact point and surface normals for the finger $i$, $M \in \bm{\mathrm{SE}(3)}$ is the wrist pose. 
Achieving static equilibrium when grasping requires that the force applied by each finger on the object stays within the friction cone \cite{nguyen1988constructing}. In this sense, we attempt to select finger positions applying minimal tangential forces on the object during grasping. This is accomplished by finding contact points maximizing the dot product between finger and object surface normal vectors. Correlating the BEGIs corresponding to object and gripper fingers 
provide indication on the location of such points in $\bm{\mathrm{SO}(3)}$. Rotations with high correlation values can then be sampled from the correlation density function, while the BEGI point set map $\mathcal{P} = \{\mathcal{P}_{jk}\}$ is used to extract, for each finger, those contact points. The following contact sampling algorithm is then devised:
\begin{itemize}
    \item Given an object point cloud and a hand configuration parametrized by the gripper's joint configuration $\bm{q}$, we compute the BEGIs of point cloud and gripper fingers using \eqref{eq:begi_representation}.
    \item The correlation between these two BEGIs is then computed using \eqref{eq:correlation_simplified}. This process is depicted in Fig. \ref{fig:begi_correlation}. 
    \item Rotations with correlation values greater than a threshold $t_{corr}$ are then sampled. The BEGI function of the gripper is then rotated and for each finger, object points located at the index $(j,k)$ of the finger are extracted.
    \item The set $\mathcal{P}^r_{N_f} = \{\mathcal{P}_{(0)}, \mathcal{P}_{(1)}\cdots \mathcal{P}_{(N_f)}\}$  can then be constructed for each rotation where $\mathcal{P}_{(i)}$ is the set of points extracted for the finger $i$.
    \item Contact point sets with surface normals $(p_1 \cdots p_{N_f},\allowbreak n_1\cdots n_{N_f})$ can then be extracted from $\mathcal{P}^r_{N_f}$. 
    \item An additional filtering step is required in order to remove the contacts that do not satisfy the force closure principle and that are out of range of the gripper kinematics.
    \item Wrist poses $\{M\}$ leading to collision-free grasps can then be sampled using the kinematics of the gripper.
\end{itemize}
%
%
\subsection{Grasp Ranking}
%
Grasp candidates obtained from the previous step need to be ranked such that the one with maximum likelihood is selected and executed on the robot. For this purpose, we use our previously developed LoCoMo metric \cite{adjigble2018model}, which provides a similarity score between gripper fingers and object at the contact points. Due to the generic nature of SpectGRASP, using alternative ranking metrics will not limit its performance. The benefits of using the LoCoMo metric are twofold; first, it  has been proven to perform well in practice and second, it provides a common basis to objectively evaluate the proposed method against the one presented in \cite{adjigble2018model}. LoCoMo grasp ranking $\mathcal{Q}$ is computed as in \eqref{eq:locomo}.
%
\begin{figure*}[!ht]
    \centering
    \includegraphics[width=\textwidth]{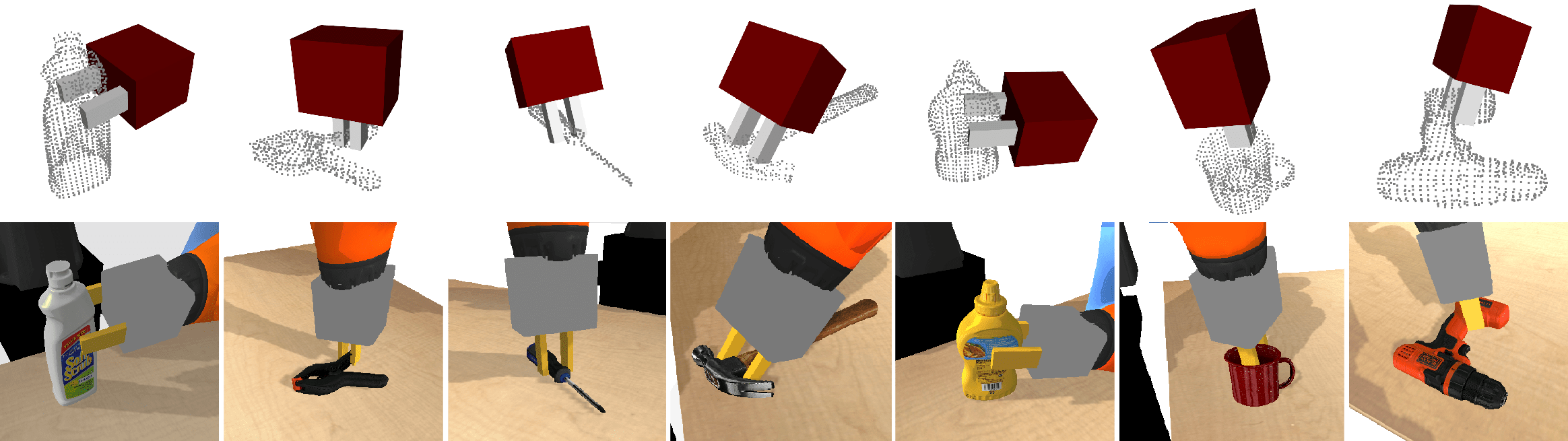}
    \caption{Successful grasps for various objects from the YCB object set using the proposed SpectGRASP method. Top row shows the computed top-ranked grasp candidates and bottom row shows the same grasps executed on the simulated robot. Detailed results can be found in the supplementary video.}
    \label{fig:single_grasps}
\end{figure*}
\begin{equation}
\label{eq:locomo}
\begin{aligned}
        \mathcal{M}_{i} &= \frac{1}{N_{s}} \sum_{j=1}^{\kappa}\left((2\pi)^{\kappa} | \Sigma| \right)^{\frac{1}{2}}\mathcal{N}\left(\varepsilon_j; \vec{0},\Sigma \right) \\ 
        \mathcal{Q} &= \rho\prod_{i=1}^{N_{f}}\mathcal{M}_{i}^{\omega_{i}}
\end{aligned}
\end{equation}
where, $\mathcal{M}_{i}$ is the contact moment (CoMo) of finger $i$ computed at a contact point,  $N_s$ is a normalizing term, $\kappa$ is the number of point cloud patches sampled around the location of the finger, $\mathcal{N}(\varepsilon_j; \vec{0},\Sigma )$ is the multivariate Gaussian density function centered at $\vec{0}$ with covariance $\Sigma$, $\varepsilon_j$ is the difference of gripper and object zero-moment shift vectors. Finally, $\omega_i$ and $\rho$ are weighting factors. As presented in \cite{adjigble2019assisted}, $\rho$ can be dynamically updated to re-rank grasps based on factors such as the distance between robot's end effector and closest grasps. More details on the derivation of  \eqref{eq:locomo} can be found in \cite{adjigble2018model}.

%% file: sections/results.tex
%
%
%
In this section, we present the experiments conducted in simulation to validate the performance of the proposed SpectGRASP method in grasping individual objects as well as clearing multi-object scenes. We also compare the grasping performance of our method with state-of-the-art LoCoMo-based grasp planner\footnote{LoCoMo grasp planner code is available at: \url{https://github.com/maximeadjigble/grasplocomo}} \cite{adjigble2018model} and the obtained results are discussed. Furthermore, we also discuss in detail the applicability of our approach to real scenarios. All experiments are performed on a windows machine with an i7 CPU.
%
%
\subsection{Simulation Setup Description}
All experiments are performed using the PyBullet physics simulator. It is a Python wrapper for the Bullet physics engine, which has been proven to be an efficient open-source simulation tool for robotic models \cite{coumans2016pybullet}. Our robotic setup consists of a 7 degrees of freedom (DoF) articulated robot arm fitted with a parallel-jaw gripper. 
For experiments, we have selected 16 different objects from the YCB object set \cite{calli2015ycb} that are suitable with our gripper model. 
In order to make our system as realistic as possible, we have created a virtual RGB-D camera for point cloud acquisition and attached it to the robot's end-effector. We use a point cloud generated by registering multiple clouds acquired by moving the robot to four different viewpoints around the objects. This registration is merely cloud stitching as the point clouds are acquired by a camera whose exact pose in the robot base frame is known \cite{marturi2019dynamic}. Object CAD models downloaded from YCB website are used for simulations. In order to reach the grasp poses, a full 6-DoF end-effector pose controller has been implemented to generate the robot trajectory. Moreover, as previously stated, kinematically infeasible configurations are discarded from the full list of generated candidates. For all experiments, an empirically selected threshold $t_{corr}=0.1$ is used to generate grasps.

We report both single and multiple objects grasping results following the evaluation protocol described in \cite{BenchmarkERL}. Similar to the setup described in \cite{BenchmarkERL}, we have considered a circular task-space with a radius of $25~\mathrm{cm}$ within the robot workspace. 
%
\midsepremove
\begin{table*}
\smaller
    \caption{Single object grasping results for both proposed and LoCoMo-based methods at test location-1.}
    \label{tab:single_test_l1}
    \centering
    \begin{threeparttable}
    \begin{tabularx}{\textwidth}{>{\hsize=0.14\hsize\raggedright\arraybackslash}X|
                              >{\hsize=0.13\hsize\centering\arraybackslash}X|
                              >{\hsize=0.08\hsize\centering\arraybackslash}X|
                             >{\hsize=0.1\hsize\centering\arraybackslash}X|
                             >{\hsize=0.1\hsize\centering\arraybackslash}X|
                             >{\hsize=0.1\hsize\centering\arraybackslash}X|
                              >{\hsize=0.13\hsize\centering\arraybackslash}X|
                              >{\hsize=0.1\hsize\centering\arraybackslash}X|
                             >{\hsize=0.1\hsize\centering\arraybackslash}X|
                             >{\hsize=0.1\hsize\centering\arraybackslash}X|
                             >{\hsize=0.1\hsize\centering\arraybackslash}X}
\toprule
    \multirow{2}{*}{\textbf{Object}} & \multicolumn{5}{c|}{\textbf{SpectGRASP}} & \multicolumn{5}{c}{\textbf{LoCoMo}} \\
    \cmidrule{2-11}
      & \textbf{Total Grasps\tnote{1} (Feasible)} & \textbf{Time [s]} & \textbf{Lift (\%)} & \textbf{Rotation (\%)} & \textbf{Shake (\%)} & \textbf{Total Grasps (Feasible)} & \textbf{Time [s]} & \textbf{Lift (\%)} & \textbf{Rotation (\%)} & \textbf{Shake (\%)} \\
     \midrule
     Cleanser & 1995 (265) & 2.55 &100 &100 &100 &7916 (379) &9.93 &100 &100 &100\\
     Cup & 790 (116) &0.42 &100 &100 &100 &3421 (90) &1.66 &100 &100 &100\\
     F. Screwdriver  &2016 (125) &0.86 &100 &100 &100 &13251 (1172) &4.60 &100 &0 &NA\tnote{2}\\
   Golf Ball  &3514 (119) &0.98 &100 &100 &100 &30479 (1684) &9.63 &100 &100 &100\\
 Hammer &3173 (537) &1.58 &100 &100 &100 &24179 (4213) &10.49 &100 &0 &NA\\
 Large Clamp &47 (12) &0.15 &100 &0 &NA &1361 (173) &1.93 &100 &0 &NA\\
 Mug  &72 (4) &0.31 &100 &0 &NA &261 (9) &4.19&100 &33 &0 \\
 P. Screwdriver  &1828 (5) &0.64 &100 &100 &100 &14281 (276) &4.7 &100 &0 &NA\\
 Potted Meat &300 (6) &0.34 &100 &100 &100 &1046 (28) &0.92 &100 &100 &100\\
 Power Drill &2410 (300) &1.48 &100 &0 &NA &28503 (1626) &17.62 &100 &100 &0\\
 Racquet Ball &614 (65) &0.53 &100 &0 &NA &16064 (2821) &10.26 &100 &100 &100\\
 Softball  &7406 (52) &2.07 &100 &100 &100 &64510 (1576) &20.25 &0 &NA &NA\\
Strawberry  &5446 (16) &1.57 &100 &100 &100 &36521 (639) &12.16 &100 &100 &0\\
Woodblock  &6001 (380) &3.33 &100 &100 &100 &13269 (782) &10.82 &100 &100 &100\\
Mustard  &4638 (440) &3.66 &100 &100 &100 &13438 (1647) &21.56 &100 &100 &100\\
     \bottomrule
\end{tabularx}
    \begin{tablenotes}
        \item[1] Total number of grasps computed and in parenthesis are the total feasible grasps after IK filtering.
        \item[2] NA -- Test not performed as previous test failed.
    \end{tablenotes}
    \end{threeparttable}
\end{table*}
%
\midsepremove
\begin{table*}
\smaller
    \caption{Single object grasping results for both proposed and LoCoMo-based methods at test location-2.}
    \label{tab:single_test_l2}
    \centering
    \begin{tabularx}{\textwidth}{>{\hsize=0.14\hsize\raggedright\arraybackslash}X|
                              >{\hsize=0.13\hsize\centering\arraybackslash}X|
                              >{\hsize=0.08\hsize\centering\arraybackslash}X|
                             >{\hsize=0.1\hsize\centering\arraybackslash}X|
                             >{\hsize=0.1\hsize\centering\arraybackslash}X|
                             >{\hsize=0.1\hsize\centering\arraybackslash}X|
                              >{\hsize=0.13\hsize\centering\arraybackslash}X|
                              >{\hsize=0.1\hsize\centering\arraybackslash}X|
                             >{\hsize=0.1\hsize\centering\arraybackslash}X|
                             >{\hsize=0.1\hsize\centering\arraybackslash}X|
                             >{\hsize=0.1\hsize\centering\arraybackslash}X}
\toprule
    \multirow{2}{*}{\textbf{Object}} & \multicolumn{5}{c|}{\textbf{SpectGRASP}} & \multicolumn{5}{c}{\textbf{LoCoMo}} \\
    \cmidrule{2-11}
      & \textbf{Total (Feasible)} & \textbf{Time} & \textbf{Lift} & \textbf{Rotation} & \textbf{Shake} & \textbf{Total (Feasible)} & \textbf{Time} & \textbf{Lift} & \textbf{Rotation} & \textbf{Shake} \\
     \midrule
  Cleanser & 146 (33) &0.32 &100 &100 &100 &8972 (1921) &10.93 &100 &100 &100\\
  Cup & 617 (9) &0.43 &100 &100 &100 &1525 (10) &1.1 &100 &100 &0\\
 F. Screwdriver & 1869 (106) &0.75 &100 &0 &NA &13210 (694) &4.65 &100 &0 &NA\\
Golf Ball & 5674 (321) &1.46 &100 &100 &100 &38481 (1891) &12.64 &100 &100 &100\\
Hammer & 2678 (31) &1.25 &100 &100 &100 &14505 (231) &7.94 &100 &100 &100\\
  Large Clamp & 161 (5) &0.24 &100 &0 &NA &1897 (69) &1.86 &100 &0 &NA\\
 Mug & 90 (14) &0.41 &0 &NA &NA &312 (38) &4.3 &100 &100 &100\\
P. Screwdriver& 1502 (228) &0.58 &100 &100 &100 &11995 (1108) &4.02 &100 &100 &100\\
 Potted Meat & 118 (8) &0.33 &100 &100 &100 &1203 (94) &1.06 &0 &NA &NA\\
 Power Drill & 2767 (577) &1.57 &100 &100 &100 &28078 (1869) &20.6 &100 &0 &NA\\
Racquet Ball & 505 (79) &0.45 &100 &100 &100 &14425 (1084) &8.64 &0 &NA &NA\\
Softball & 8489 (327) &2.45 &100 &100 &100 &68615 (4286) &21.58 &100 &0 &NA\\
Strawberry & 5164 (419) &1.54 &100 &100 &100 &34740 (1613) &11.47 &0 &NA &NA\\
Woodblock & 5152 (707) &2.93 &100 &100 &100 &12262 (1155) &11.87 &100 &100 &100\\
Mustard   & 3428 (928) &3.35 &100 &100 &100 &13080 (1865) &21.68 &100 &100 &100\\
     \bottomrule
    \end{tabularx}
    \vspace{-5pt}
\end{table*}
\subsection{Grasping Individual Objects}
For this experiment, two different test locations on the circular task-space are selected. For location-1, the objects are placed at the origin of the circular task-space, \textit{i.e.}, the point on the ground plane at which the centre of the tool is projected onto, when the robot is at 90-90 configuration. Refer \cite{BenchmarkERL} for more details. For location-2, we move the object $25~\mathrm{cm}$ in the negative-$y$ direction and apply a rotation of $-90^{\circ}$ around $z$-axis. After candidate generation, filtering, and sequential ranking, the top ranked grasp is executed on the robot. A force of $50\mathrm{N}$ is used to grasp and hold the object. Thereafter, a series of three tests, as in \cite{BenchmarkERL}, are performed to check the grasp stability: (i) lift test, where the object is lifted $20~\mathrm{cm}$ off the table at a speed of $10~\mathrm{cm/s}$; (ii) rotation test, where the object is rotated $90^{\circ}$ and $-90^{\circ}$ around the $y$-axis at a speed of $45~\mathrm{deg/s}$; and finally (iii) shake test, where the robot shakes the object in a sinusoidal pattern ($0.25 \mathrm{m}$ amplitude, $10~\mathrm{m/s^2}$ acceleration) for 10 seconds. These tests are performed sequentially, \textit{i.e.}, if any of these fail the grasp is considered a failure and the next test will not be performed. The test is considered a fail if the object slips out of the fingers. For each object, with both proposed and LoCoMo-based methods, at each test location, we have repeated the test 3 times. In the paper, we report the average of those three trials.%

Fig.~\ref{fig:single_grasps} shows some screenshots of the successful grasps generated and executed for 7 different objects. Tables \ref{tab:single_test_l1} and \ref{tab:single_test_l2} summarise the results for test locations 1 and 2, respectively. From the results, it can be seen that the proposed method outperformed the LoCoMo-based grasp planner in terms of grasp generation time. SpectGRASP takes on average $\bm{1.28~\mathrm{seconds}}$ to generate grasps. This is $\bm{7\times}$ faster compared to the LoCoMo-based grasp planner, which takes $9.5~\mathrm{seconds}$ on average. Although SpectGRASP generates fewer grasp candidates compared to LoCoMo, it demonstrated higher success rates than the later. For location 1,  the average success rates of SpectGRASP for lifting, rotation and shaking tests are respectively, $\bm{100\%}$, $\bm{73\%}$ and $\bm{73\%}$, while for LoCoMo they are $93\%$, $62\%$, $47\%$. And for location 2, the average success rates of SpectGRASP for lifting, rotation and shaking tests are respectively $\bm{93\%}$, $\bm{80\%}$ and $\bm{80\%}$, while for LoCoMo they are $80\%$, $53\%$, $47\%$. These results suggest that the SpectGRASP is able to generate good and stable grasps than the ones generated by LoCoMo-based method.

The difference in the number of grasps generated is due to the fact that LoCoMo uses all possible point pairs of a given point cloud to sample contact points. This results in a high number of grasp candidates; however, the generation time is more. Nevertheless, in practical real-world applications, only the top ranked grasps are considered; thus, making the SpectGRASP a more suitable grasp planner. It is worth noting that for both methods, the success rates from rotation and shake tests are lower than that of the lifting test. This is mainly due to the inaccuracy of physics simulator to model interactions between objects. We believe that these success rates will be improved with a real robot system, e.g. see results for LoCoMo method in \cite{adjigble2018model} and \cite{BenchmarkERL}. 
\begin{figure*}
    \centering
    \includegraphics[width=\textwidth]{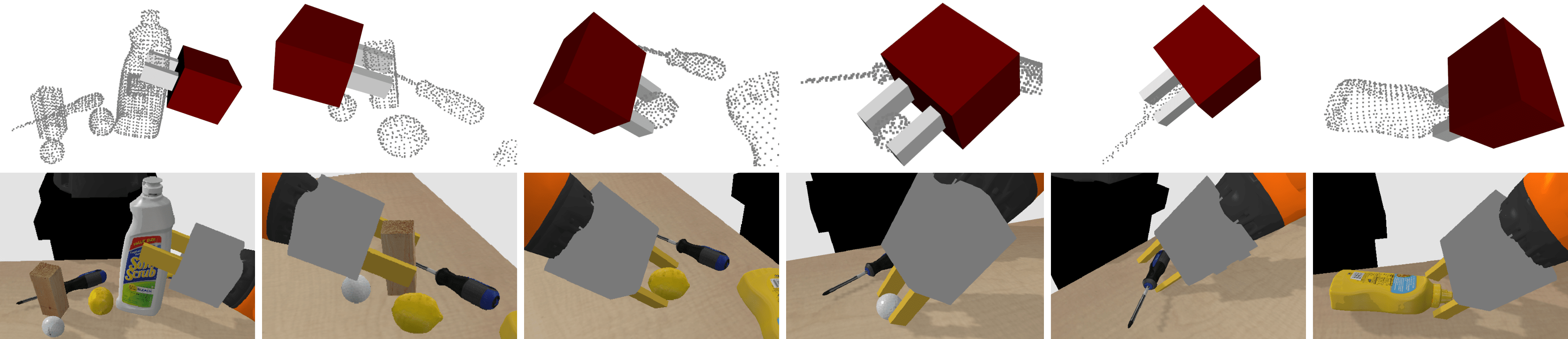}
    \caption{Grasping objects in clutter with proposed method. Top row shows generated top-ranked grasps and the bottom row shows the same grasps executed on the robot. Objects are grasped one by one until either the scene is cleared, or the method returns no feasible grasps. Detailed results can be found in the supplementary video.}
    \label{fig:clutter_grasps}
\end{figure*}
\subsection{Grasping Objects in Clutter}
The second set of experiments are performed to evaluate the performance of the proposed method in clearing cluttered scenes containing multiple objects. For this tests, 6 different objects are randomly positioned. We test the performance by clearing 3 such randomly generated scenes.  The task is to successively grasp, lift, rotate and shake one object at a time until the scene is successfully cleared. For each iteration, a new scene point cloud is captured based on which the grasps are generated. Like before, the generated grasps are filtered and ranked, and the top grasp is executed by the robot. The experiment is repeated until all objects are successfully removed or if the algorithm fails to find any feasible grasp. Also for this tests, we report the results following the protocol in \cite{BenchmarkERL} for group of objects. 

Fig. \ref{fig:clutter_grasps} shows successive images from this test where a cluttered scene being cleared using the proposed method. Table \ref{tab:clutter_table} summarises the result for clearing three different clutters. SpectGRASP is able to successfully clear all 3 scenes. It can be seen that the pickup order varies due to the random positioning of objects, but in all cases the first 4 objects are successfully picked up in the first attempt. For the 6\textsuperscript{th} object of scene 2, the ``Mustard'' bottle was lifted successfully but fell back on the table during the rotation test. Another attempt was required to successfully complete this trial. Likewise, two attempts were needed to clear ``screwdriver'' in scene 3. Overall, considering all trials, on average it took $18.02~\mathrm{seconds}$ to generate grasps required to clear a scene of 6 objects. This corresponds to $ \approx3~\mathrm{seconds}$ per object, which is slightly higher compared to the case of single object grasping. This is because, cluttered scenes are composed of multiple objects, which means more grasp candidates are generated leading to high generation time.
%
\setlength{\textfloatsep}{5pt }
\midsepremove
\begin{table}
\smaller
    \caption{Clutter clearance results with proposed method.}
    \label{tab:clutter_table}
    \centering
    \begin{threeparttable}
    \begin{tabularx}{\columnwidth}{>{\hsize=0.07\hsize\centering\arraybackslash}X|
                              >{\hsize=0.18\hsize\centering\arraybackslash}X|
                              >{\hsize=0.1\hsize\centering\arraybackslash}X|
                             >{\hsize=0.1\hsize\centering\arraybackslash}X|
                             >{\hsize=0.07\hsize\centering\arraybackslash}X|
                             >{\hsize=0.1\hsize\centering\arraybackslash}X|
                              >{\hsize=0.07\hsize\centering\arraybackslash}X}
\toprule
    Scene & Object & Pickup order & Time (s) & Lift & Rotation & Shake\\
    \midrule
    \multirow{6}{*}{\textbf{1}} & Woodblock& 2 & 3.96 &\mytick &\mytick &\mytick \\
        & Golfball & 4 & 2.37 & \mytick & \mytick & \mytick \\
        & Dishwash & 1  & 7 & \mytick & \mytick & \mytick \\
        & Lemon & 3 & 2.67 & \mytick & \mytick & \mytick \\
        & Screwdriver& 5 & 0.75 & \mytick & \mytick & \mytick \\
        & Mustard& 6 & 0.14 & \mytick & \mytick & \mytick \\
    \midrule
    \multicolumn{3}{r}{Total grasp generation time: }& \multicolumn{4}{l}{16.89 seconds}\\
    \midrule
     \multirow{6}{*}{\textbf{2}} & Woodblock & 1 & 5.18 & \mytick & \mytick & \mytick \\
        & Golfball & 3 & 2.11 & \mytick & \mytick & \mytick \\
        & Dishwash & 2 & 3.4 & \mytick & \mytick & \mytick \\
        & Lemon & 4 & 1.09 & \mytick & \mytick & \mytick  \\
        & Screwdriver & 5 & 0.63 & \mytick & \mytick & \mytick  \\
        & Mustard & 6 & 0.25 & \mytickb \tnote{3} & \mytickb & \mytickb  \\
    \midrule
    \multicolumn{3}{r}{Total grasp generation time: }& \multicolumn{4}{l}{12.66 seconds}\\
    \midrule
     \multirow{6}{*}{\textbf{3}} & Woodblock & 2 & 5.59 & \mytick & \mytick &\mytick \\
        & Golfball & 6 & 1.66 &\mytick &\mytick &\mytick  \\
        & Dishwash & 1 & 6.45 & \mytick & \mytick & \mytick  \\
        & Lemon & 4 & 3.34 &\mytick &\mytick &\mytick  \\
        & Screwdriver & 5 & 2.96 & \mytickb & \mytickb & \mytickb  \\
        & Mustard & 3 & 4.51 &\mytick &\mytick & \mytick \\
    \midrule
    \multicolumn{3}{r}{Total grasp generation time: }& \multicolumn{4}{l}{24.51 seconds}\\
    \midrule
    \end{tabularx}
    \begin{tablenotes}
        \item[3] Purple tick mark: Success after second try.
    \end{tablenotes}
    \end{threeparttable}
\end{table}
\subsection{Discussion}\label{sec:discussions}
\subsubsection*{\bfseries Complexity analysis} 
Experimental results demonstrate the efficiency of SpectGRASP method in generating grasps for various objects with a variety of shapes. This is mainly supported by rapid computation of the correlation density map that is used to sample contact points. For a parallel jaw gripper, naively sampling contact points on a point cloud with $N$ points results in a $\mathcal{O}(N^2)$ complexity. As the number of points increases, the cost of sampling contacts becomes high. Also, this complexity drastically increases when hands with more fingers are used. 
However, for SpectGRASP, which uses the Fourier transform on $\bm{\mathrm{SO}(3)}$, has a complexity of $\mathcal{O}(B^4)$ with $B$ being the bandwidth \cite{kostelec2008ffts}. %
Suppose, $N_r$ rotations are sampled from the correlation density map and a maximum number of $K$ points are extracted from each BEGI cell, in extreme cases, sampling contacts leads to an $\mathcal{O}(N_rK^{N_f})$ complexity, with $N_f$ being the number of fingers. It is important to notice that the complexity of SpectGRASP is not directly related to the number of points in the point cloud, meaning that the efficiency of the method still remains same even for dense point clouds. Additionally, $B$ value can be adjusted to modify the grasp generation time. For instance, lower values of $B$  lead to higher number of points per cell but result in a lower number of possible rotations. This is desirable for applications requiring high number of grasp candidates. While, higher values of B tend to produce a smaller number of points per cell with a high number of rotations. This leads to fewer grasp candidates with precise matching. Applications such as dynamic grasping could greatly benefit from this type of fine matching. The contact sampling complexity can be further reduced by clustering the points in each BEGI cell. 

\subsubsection*{\bfseries Applicability to real system} In this work, we have validated our approach by a substantial number of virtual experiments with a simulated 7-DoF robot and a parallel-jaw gripper, grasping a variety of objects. The selected robot and gripper configurations match with the real KUKA iiwa robot arm and Schunk PG-70 gripper models, same as the setup presented in \cite{adjigble2018model}. The full pipeline presented in this paper can be directly imported on to the real robot as it does not assume any kinematic constraints, and the grasp configurations for which there is no IK solutions are discarded. Our method relies only on the point clouds generated by the simulated virtual camera, which is mounted on the robot’s wrist. This camera has been constructed considering the camera parameters of a real depth camera. This means that the virtually generated point clouds are similar to that of the real-world depth sensor. As previously stated, our autonomous grasping technique does not rely on any prior knowledge of the object, e.g., a CAD model. Furthermore, our approach does not make use of any additional information regarding given unknown object that would not be available in a real-world scenario. The assumption is that the object will be placed in a reachable location by the robot within its workspace, viewable from a camera and the object size should be suitable for the gripper to be able to manipulate. Besides, our approach is agnostic to the robot system used, given that: (i) there is a scene or robot mounted vision system to acquire point cloud; (ii) the robot’s end-effector is a multi-finger hand with minimum two fingers; and (iii) hand kinematic model is known. Henceforth, we believe that our method is applicable to real systems. 

%% file: sections/conclusion.tex
In this paper, we have presented a computationally efficient and robust method, SpectGRASP, for robotic grasping via spectral correlation and grasp ranking. A Fast Fourier Transform of functions defined on the sphere is used to efficiently sample contact points on the surface of the object, while the LoCoMo metric is used for grasp ranking. The method uses only the point cloud and surface normals of an object to generate grasps. No training data, learning nor any 3D object model is needed. We also introduced a novel representation of point clouds with surface normals (BEGI). This can be used to efficiently find correspondance points between two point clouds. Experimental results in simulation have shown the efficiency and robustness of SpectGRASP for grasp generation using a parallel-jaw gripper. SpectGRASP outperforms the previous state-of-the-art ``model-free and learning-free grasping'' method, in terms of grasp generation time and success rate for the tasks of grasping single objects and clearing cluttered scenes with successive grasps.%

Future work will focus on evaluating the method on a real robot setup, and also on
extending it to multi-finger hands. Due to the swiftness and efficiency of the grasp generation pipeline, we intend to use it for tracking and grasping moving objects by integrating it with the method presented in \cite{defarias2021dual}.